\documentclass{article}

\PassOptionsToPackage{numbers, compress}{natbib}
 \usepackage[preprint]{neurips_2026}

\usepackage[utf8]{inputenc} 
\usepackage[T1]{fontenc}    
\usepackage{hyperref}       
\usepackage{url}            
\usepackage{booktabs}       
\usepackage{amsfonts}       
\usepackage{nicefrac}       
\usepackage{microtype}      
\usepackage{xcolor}         
\usepackage[table]{xcolor}
\usepackage{amsmath}
\usepackage{graphicx}
\usepackage{subcaption}
\usepackage{algorithm}
\usepackage{algpseudocode}
\usepackage{wrapfig}
\usepackage{multirow}

\title{Edit-GRPO: A Locality-Preserving Policy Optimization Framework for Image Editing}

\author{
\textbf{Shaodong Xu} \quad
\textbf{Zexian Li} \quad
\textbf{Zhendong Wang} \quad
\textbf{Litong Gong} \quad
\textbf{Tiezheng Ge} \\
\textbf{Wengang Zhou} \quad
\textbf{Bo Zheng} \quad
\textbf{Houqiang Li} \\
Alibaba Group \\
}

\begin{document}

\maketitle

\begin{abstract}
A fundamental challenge in image editing lies in preserving spatial locality: edits should improve targeted content without inadvertently altering surrounding regions. However, most optimization-based editing approaches treat images as holistic entities, causing global policy updates that undermine locality and introduce undesired context changes. We observe that this issue stems from a mismatch between localized editing intent and globally applied optimization signals. Motivated by this insight, we propose Edit-GRPO, preserving Locality while optimizing image editing, a locality-preserving policy optimization framework that explicitly decouples editing and preservation objectives. By assigning region-specific optimization signals to edit and non-edit areas, Edit-GRPO aligns policy updates with the spatial structure of editing tasks, enabling localized improvements while maintaining global visual coherence. This design effectively suppresses common artifacts such as context distortion and boundary inconsistency. Extensive experiments across diverse image editing scenarios demonstrate that Edit-GRPO significantly improves locality preservation while maintaining strong editing performance compared to existing optimization-based methods, validating the generality and effectiveness of the proposed framework. 
\end{abstract}

\section{Introduction}
\label{sec:intro}
Recent advances in generative models, particularly diffusion models~\cite{ho2020denoising, rombach2022high, Peebles2022ScalableDM} and flow models~\cite{lipman2022flow, liu2022flow, esser2024scaling} have significantly improved visual content creation, enabling high-quality instruction-guided image editing~\cite{labs2025flux, wu2025qwen}, which requires both precise semantic control and faithful context preservation.

Inspired by the success of reinforcement learning from human feedback (RLHF) in LLMs~\cite{ouyang2022training, guo2025deepseek, shao2024deepseekmath}, recent work has explored integrating RL with visual generative models~\cite{black2023training, Fan2023DPOKRL}. Recently, one line of work~\cite{liu2025flow, xue2025dancegrpo, li2025mixgrpo} introduces GRPO-style policy optimization to diffusion models by converting the ODE sampling into an equivalent SDE. Another line of work~\cite{zheng2025diffusionnft,li2025uniworld} performs policy optimization directly on the forward diffusion process via flow matching objectives, enabling more efficient training. Despite their differences in policy optimization, these RL-based approaches share a common design: \textit{reward signals are computed at image level and used to guide policy updates globally}.

While this global optimization paradigm works well for general image generation, it becomes problematic when applied to image editing. Unlike generation, editing is inherently a localized task: \textbf{only the target regions should be modified, while the remaining content should remain unchanged}. However, we observe that existing optimization-based approaches~\cite{liu2025flow, xue2025dancegrpo, li2025mixgrpo, zheng2025diffusionnft,li2025uniworld} typically treat images as holistic entities and use image-level rewards to update the policy, which conflicts with the localized nature of editing. More importantly, a single global reward can \textbf{not} explicitly distinguish where the model should edit and where it should preserve, making the editing model struggle to achieve precise semantic modification and faithful context preservation simultaneously. When the model is encouraged to perform stronger semantic edits, unintended distortion beyond the editing areas arises inadvertently (such as background drift and boundary inconsistency), degrading the final editing quality. This observation suggests that, in image editing, \textbf{locality preservation should not be treated as a byproduct of semantic editing, but as an explicit optimization objective during post-training}. Consequently, reward signals should be region-specific rather than globally defined: semantic following reward should be evaluated in the target regions, while locality preservation reward should be evaluated individually in the non-edit regions. This insight motivates us to explore a decoupled reward mechanism that assesses semantic editing and locality preservation separately, which remains underexplored in image editing post-training.

In this work, we propose \textbf{Edit-GRPO}, a policy optimization framework that \textbf{preserves locality while optimizing image editing}. The core idea is to decouple the reward design based on the spatial regions of the edited images. Specifically, we partition the images into edit regions and non-edit regions, and design different rewards separately. For edit regions, we apply a semantic following reward to encourage precise editing that follows the instruction. Meanwhile, we introduce a locality preservation reward for the non-edit regions that explicitly encourages the model to retain the original visual content in areas unrelated to the edit. Based on the decoupled reward mechanism, we further propose a region-decoupled optimization framework to mitigate the interference between editing and preservation objectives. By decoupling optimization objectives in this way, Edit-GRPO enables joint policy optimization for improving both semantic editing and locality preservation, encouraging the model to produce more visually coherent and realistic editing results.

To validate Edit-GRPO, we apply our method to FLUX.1-Kontext [Dev]~\cite{labs2025flux} and Qwen-Image-Edit [2509]~\cite{wu2025qwen}. Extensive experimental results demonstrate that Edit-GRPO consistently improves locality preservation, effectively suppressing unintended changes in non-edit regions.
Importantly, Edit-GRPO achieves comparable or even better semantic editing performance relative to the base models, indicating that the locality-preserving improvements do not compromise edit capability. Together, these results show that Edit-GRPO improves editing precision and locality preservation simultaneously, highlighting the effectiveness of the proposed framework.

\begin{figure}[t]
    \centering
    \includegraphics[width=\linewidth]{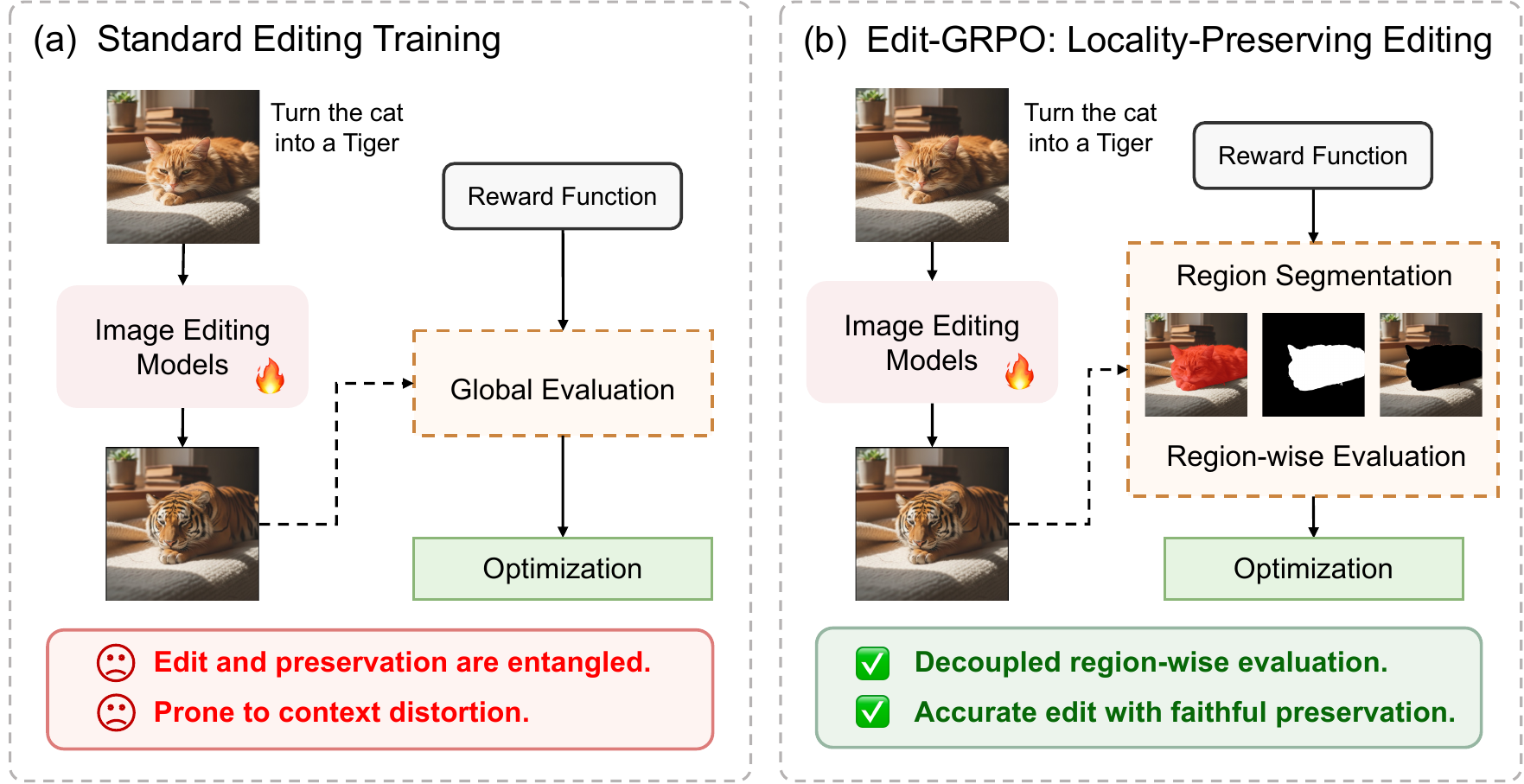}
    \caption{
    Overview of the proposed Edit-GRPO framework compared to standard editing post-training. Edit-GRPO performs region-wise evaluation to better preserve non-edit content.}
    \vspace{-5pt}
    \label{fig:framework}
\end{figure}

The main contributions of this work are summarized as follows:
\begin{itemize}
    \item We identify a fundamental mismatch between localized editing intent and globally applied optimization signals in existing optimization-based image editing methods.
    \item We propose Edit-GRPO, a locality-preserving policy optimization framework that decouples editing and preservation objectives through region-specific optimization signals.
    \item Extensive experiments demonstrate that Edit-GRPO consistently improves both editing fidelity and context preservation over the base models.
\end{itemize}

\section{Related Work}
\subsection{Instruction-Based Image Editing}
Diffusion and flow models~\cite{ho2020denoising, Song2020DenoisingDI, rombach2022high,Song2020ScoreBasedGM, lipman2022flow, liu2022flow} have significantly advanced text-to-image generation and instruction-based image editing, allowing users to manipulate visual content through natural language instructions. Early works~\cite{Meng2021SDEditGI, brooks2022instructpix2pix, Wang2023InstructEditIA, Zhang2023MagicBrushAM} demonstrate that diffusion models can acquire instruction-following editing capabilities through supervised fine-tuning. Recent works~\cite{labs2025flux, wu2025qwen, liu2025step1x, Cai2025HiDreamI1AH, Deng2025EmergingPI} further improve editing performance by scaling model capacity and training data. However, these Supervised Fine-Tuning (SFT) approaches often lead to shortcut learning and overfitting to large-scale datasets~\cite{li2025uniworld}, limiting the generalization and controllability of editing models. To this end, post-training alignment methods based on reinforcement learning (RL) have emerged as a promising solution.

\subsection{Policy Optimization for Generative Models}
Inspired by the success of reinforcement learning from human feedback in LLMs \cite{ouyang2022training, guo2025deepseek, shao2024deepseekmath}, recent efforts have extended RL to visual generative models. Pioneer works like DDPO~\cite{black2023training} and ReFL~\cite{Xu2023ImageRewardLA} apply PPO~\cite{Schulman2017ProximalPO} to fine-tune diffusion models for human preference alignment. Diffusion-DPO~\cite{Wallace2023DiffusionMA} adapts Direct Preference Optimization (DPO)~\cite{Rafailov2023DirectPO} to directly optimize diffusion models from paired data, bypassing the need for explicit reward models.
Subsequent studies such as Flow-GRPO~\cite{liu2025flow} and Dance-GRPO~\cite{xue2025dancegrpo} enable GRPO-style policy updates by converting the ODE sampling into an equivalent SDE. MixGRPO~\cite{li2025mixgrpo} further improves training efficiency through a hybrid ODE-SDE sampling approach. More recently, DiffusionNFT~\cite{zheng2025diffusionnft, li2025uniworld} directly optimizes diffusion models along the forward process via flow matching objectives, leading to better generation performance. Despite these advances, existing optimization methods typically treat images as holistic entities and rely on image-level rewards to update the policy, which works well for generic image generation but becomes problematic when applied to image editing, where edits are expected to occur only in specific regions without altering surrounding content. Achieving the balance between precise semantic editing and faithful locality preservation remains an open challenge in complex image editing tasks.

\section{Preliminaries}
In this section, we briefly review the formulation of flow matching~\cite{lipman2022flow, liu2022flow} and Flow-GRPO~\cite{liu2025flow}.

\textbf{Flow Matching.}
Let $x_0 \sim X_0$ denote a clean image sample and $x_1 \sim X_1$ denote a Gaussian noise sample. Following rectified flow, the interpolated sample at time $t \in [0,1]$ is defined as:
\begin{equation}
    x_t = (1 - t)x_0 + tx_1,
\end{equation}
with target velocity field $v = x_1 - x_0$.
A neural network $v_\theta(x_t, t, c)$ conditioned on instruction $c$ is trained to regress the target velocity by minimizing the flow matching objective:
\begin{equation}
    \mathcal{L}_{\mathrm{FM}}(\theta)=\mathbb{E}_{t, x_0 \sim X_0, x_1 \sim X_1}\left[\| v - v_\theta(x_t, t, c) \|_2^2\right].
\end{equation}
At inference time, edited images are generated by solving the deterministic ODE conditioned on the source image $I$ and editing instruction $c$.

\textbf{Flow-GRPO.}
Our approach is built upon Flow-GRPO, which formulates the denoising process as a Markov Decision Process. Specifically, given a prompt $c$, the flow model $\pi_\theta$ first samples a group of $G$ images $\{x_0^i\}_{i=1}^G$ and their corresponding reverse-time trajectories $\{(x_T^i, x_{T-1}^i, \dots, x_0^i)\}_{i=1}^G$. Then the group-relative advantage $\hat{A}_t^i$ is computed by normalizing the rewards within the group:
\begin{equation}
    \hat{A}_t^i = \frac{R(x_0^i, c) - \text{mean}(\{R(x_0^j, c)\}_{j=1}^G)}{\text{std}(\{R(x_0^j, c)\}_{j=1}^G)}.
\end{equation}
The policy is optimized by maximizing the following objective:
\begin{equation}
\begin{split}
    \mathcal{J}_{\text{Flow-GRPO}}(\theta) = \mathbb{E} \Big[ \frac{1}{G} \sum_{i=1}^G \frac{1}{T} \sum_{t=0}^{T-1} \min \Big(r_t^i(\theta) \hat{A}_t^i, \text{clip}(r_t^i(\theta), 1\pm\epsilon) \hat{A}_t^i \Big) - \beta D_{\text{KL}}(\pi_\theta \| \pi_{\text{ref}}) \Big],
\end{split}
\end{equation}
where $r_t^i(\theta) = \frac{p_\theta(x_{t-1}^i | x_t^i, c)}{p_{\theta_{\text{old}}}(x_{t-1}^i | x_t^i, c)}$ denotes the probability ratio. 

\section{Method}
\label{sec:method}
\begin{figure*}[t]
\centering
\includegraphics[width=\linewidth]{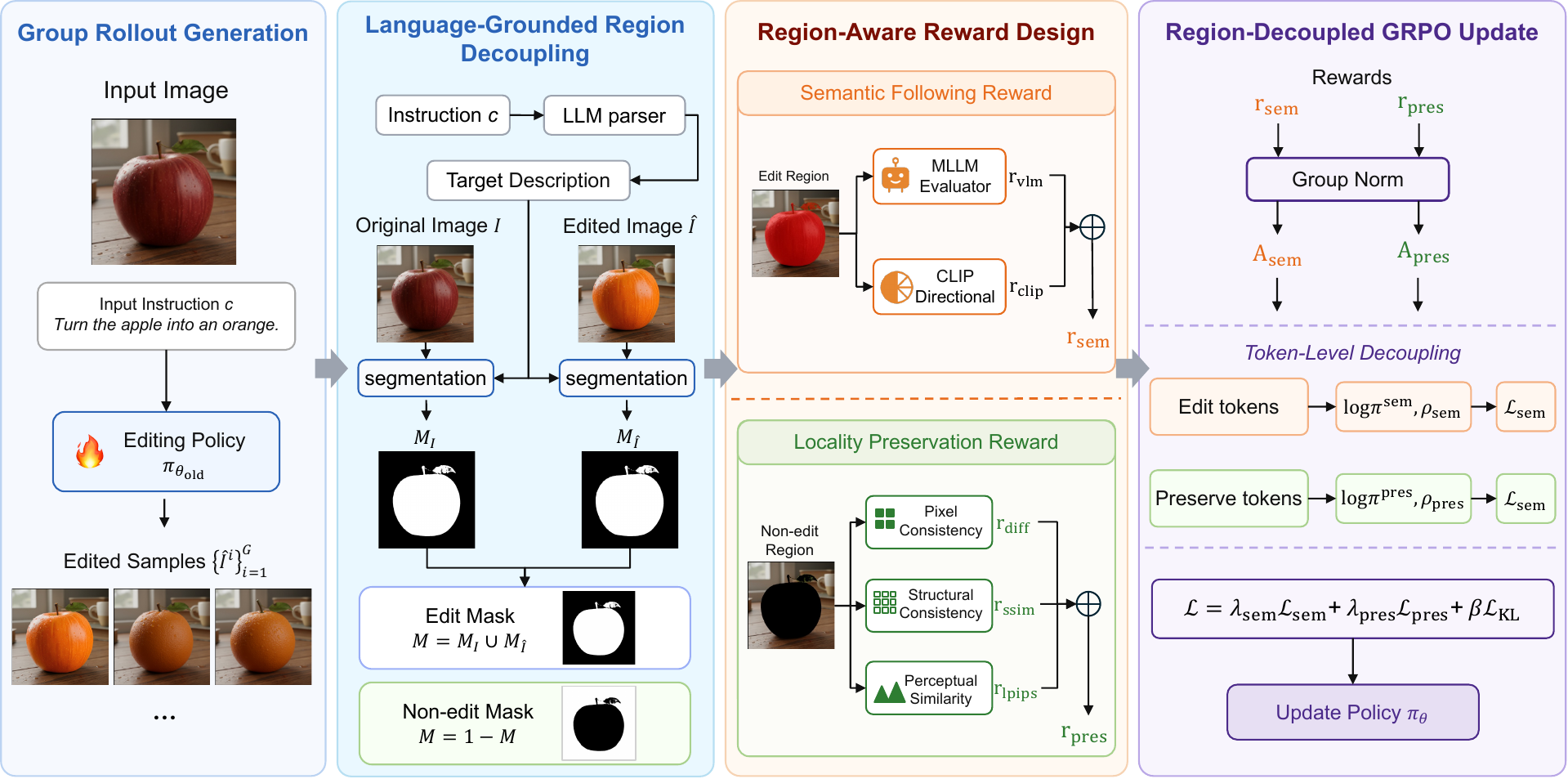}
\caption{Overview of the proposed Edit-GRPO pipeline.}
\label{fig:pipeline}
\end{figure*}

In this section, we present Edit-GRPO, a novel policy optimization framework that preserves locality while optimizing image editing. We begin by analyzing the limitations of existing optimization-based image editing approaches in Section~\ref{subsec:locality-mismatch}; then we show how to identify and decompose edit regions and design region-specific rewards in Section~\ref{subsec:segmentation} and Section~\ref{subsec:reward-design}; finally, in Section~\ref{subsec:optimization-framework} we propose a region-decoupled training framework. The overall pipeline of Edit-GRPO is illustrated in Figure~\ref{fig:pipeline}.

\subsection{Mismatch Between Localized Editing and Global Optimization Signal}
\label{subsec:locality-mismatch}
\begin{wrapfigure}{r}{0.52\linewidth}
\vspace{-14pt}
\centering
\includegraphics[width=\linewidth]{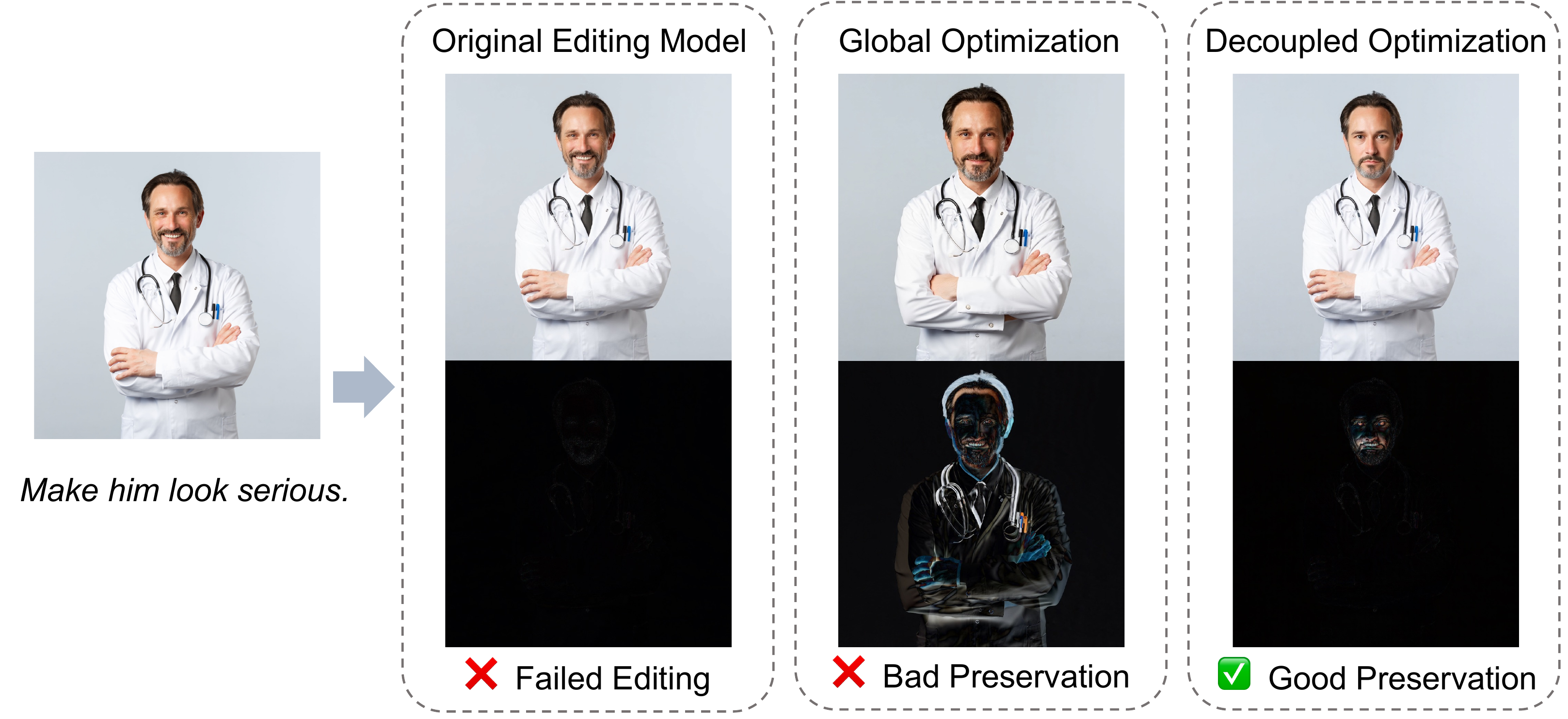}
\caption{Visual comparison between global and decoupled optimization. Please zoom in for details.}
\label{fig:motivation}
\vspace{-10pt}
\end{wrapfigure}
Image editing is fundamentally different from general image generation, as it requires not only \textit{precise semantic control} over the target content, but also \textit{faithful context preservation} of the background.
However, as illustrated in Figure~\ref{fig:framework}, existing optimization-based approaches treat images as holistic entities and conduct global evaluation over the whole image, omitting the inherently localized nature of editing. While such a \textit{holistic} optimization strategy works well for image generation, it becomes problematic when applied to image editing tasks. As shown in Figure~\ref{fig:motivation}, while global optimization successfully encourages the model to follow the editing instruction, it induces undesired background distortions, leading to a suboptimal editing result. This phenomenon reveals a fundamental mismatch between localized editing intent and globally applied optimization signals, as global optimization cannot precisely localize the editing region, causing gradients to inevitably affect the background pixels.

Motivated by this observation, we argue that \textbf{reward design in image editing should be region-specific rather than globally defined}. To this end, we propose a novel locality-preserving policy optimization framework that separates the preservation objective from the editing one. By assigning region-specific reward signals to the edit and non-edit regions, the model can improve semantic following within the target area while penalizing undesired modifications elsewhere. As shown in Figure~\ref{fig:motivation}, the region-decoupled optimization strategy better preserves surrounding context while achieving the intended edit, resulting in more faithful and spatially consistent image editing.

\subsection{Language-Grounded Region Decoupling Module}
\label{subsec:segmentation}
To achieve accurate decoupling for the edit and non-edit regions without manual annotations, we propose a Language-Grounded Region Decoupling Module that bridges free-form editing instructions and spatial segmentation of edited images.
Specifically, given an input image $I$ and an editing instruction $c$, we first use an LLM to parse the instruction and extract structured descriptions of the target content. These descriptions and images are then fed into a segmentation model to generate high-resolution segmentation masks for the edit target. To acquire robust segmentation in diverse editing scenarios, we apply region decoupling to the original image $I$ and the edited image $\hat{I}$, obtaining two individual masks. The final mask $M$ is computed as the union of the two masks:
\begin{equation}
    M =\mathcal{S}(I, c) \cup\mathcal{S}(\hat{I}, c),
\end{equation}
where $\mathcal{S}$ denotes the language-grounded region decoupling module. The remaining areas are defined as the non-edit areas $\bar{M}=1-M$. By grounding editing instructions into precise spatial masks, our approach eliminates the need for expensive user annotations, automatically generating precise masks for the edit regions in instruction-based image editing tasks.

\subsection{Region-Specific Reward Design}
\label{subsec:reward-design}
Based on the segmentation masks, we further design a region-specific reward mechanism that decomposes the optimization objective into two distinct components: a \textbf{semantic following reward}, which evaluates the precision of the intended edit, and a \textbf{locality preservation reward}, which measures the preservation of the non-edit background. We detail our reward design as follows:

\textbf{Semantic Following Reward.} 
This reward assesses how accurately the edited image reflects the semantic intent of the editing instruction.
Recent studies~\cite{wang2025unified, xu2026visionreward, gong2025onereward, wu2025editreward, li2025uniworld} have revealed that MLLMs demonstrate superior capability in evaluating image quality due to their strong visual understanding and human preference alignment. In Edit-GRPO, we leverage a pretrained MLLM as a training-free reward model to evaluate editing quality. Specifically, given the original image $I$, the edited image $\hat{I}$ and the editing instruction $c$, instead of using discrete output scores, we utilize \textbf{logit-based scores} calculated from the score token probability to obtain fine-grained continuous reward signal~\cite{Wu2025RewardDanceRS, Zhang2025UPMEAU, li2025uniworld}. Formally, we define the MLLM reward as:
\begin{equation}
    r_{\text{vlm}} = \sum_{s \in \mathcal{M}} s \cdot p(s | I, \hat{I}, c),
    \label{eq:vlm_reward}
\end{equation}
where $\mathcal{M}$ is the set of predefined numerical score tokens, $s$ is the numerical value of a score token, and $p(s | \cdot)$ is the softmax probability of that token. In addition, to ensure the edit aligns well with the instruction, we further incorporate a CLIP-based directional reward:
\begin{equation}
    r_{\text{clip}} = \cos\!\left(f_{\text{CLIP}}(\hat{I}) - f_{\text{CLIP}}(I),f_{\text{CLIP}}(c)\right).
\end{equation} 
The final semantic following reward is defined as the weighted sum of the two rewards above:
\begin{equation}
r_{\text{sem}} = \lambda_{\text{vlm}} \, r_{\text{vlm}} + \lambda_{\text{clip}} \, r_{\text{clip}}.
\end{equation}
where $\lambda_{\text{vlm}}$ and $\lambda_{\text{clip}}$ are coefficients balancing the contributions of the MLLM-based semantic reward and the CLIP-based directional reward, respectively.

\textbf{Locality Preservation Reward.}
To alleviate unintended changes outside the target regions, we design a locality preservation reward based on the non-edit regions to measure visual consistency. Specifically, we propose a reconstruction-based reward to capture pixel-level consistency between original image $I$ and edited image $\hat{I}$:
\begin{equation}
r_{\text{diff}} = 1 - \operatorname{L1}\left(I \odot \bar{M}, \hat{I} \odot \bar{M}\right),
\end{equation}
This term enforces strict pixel-level consistency in the non-edit regions. Additionally, to preserve structural consistency and perceptual similarity of the edited images, we further introduce a SSIM-based structural reward and a LPIPS-based perceptual reward:
\begin{equation}
r_{\text{ssim}} = \operatorname{SSIM}\!\left(I \odot \bar{M},\; \hat{I} \odot \bar{M}\right), \
r_{\text{lpips}} = -\operatorname{LPIPS}\!\left(I \odot \bar{M},\; \hat{I} \odot \bar{M}\right)
\end{equation}
The overall locality preservation reward in Edit-GRPO is formulated as the weighted sum of pixel-level reward $r_{\text{diff}}$, structural reward $r_{\text{ssim}}$ and perceptual reward $r_{\text{lpips}}$:
\begin{equation}
r_{\text{pres}} = \lambda_{\text{diff}} \, r_{\text{diff}} + \lambda_{\text{ssim}} \, r_{\text{ssim}} + \lambda_{\text{lpips}} \, r_{\text{lpips}},
\end{equation}
The scalar $\lambda_{\text{diff}}$, $\lambda_{\text{ssim}}$ and $\lambda_{\text{lpips}}$ govern the relative importance of the three preservation rewards.

\begin{algorithm}[t]
\caption{Region-Decoupled GRPO Training (\textbf{Edit-GRPO}).}
\label{alg:edit-grpo}
\textbf{Require:} Editing policy $\pi_\theta$, dataset $\{(I,c)\}$, region inference module $\mathcal{S}$, semantic evaluator $\mathcal{E}_{\text{sem}}$, preservation evaluator $\mathcal{E}_{\text{pres}}$, group size $G$.\\
\textbf{Initialize:} Reference policy $\pi_{\theta_{\text{old}}} \leftarrow \pi_\theta$.
\begin{algorithmic}[1]
\For{\text{each training iteration}}
    \For{\text{each sampled pair $(I,c)$}} \Comment{Rollout Step}
        \State Sample edited images $\{\hat{I}^i\}_{i=1}^{G} \sim \pi_{\theta_{\text{old}}}(\cdot|I,c)$.
        \State $M^i \leftarrow \mathcal{S}(I,c)\cup\mathcal{S}(\hat{I}^i,c)$.
        \State $r_{\text{sem}}^{\,i} \leftarrow \mathcal{E}_{\text{sem}}(\hat{I}^i,c,M^i)$,\quad
              $r_{\text{pres}}^{\,i} \leftarrow \mathcal{E}_{\text{pres}}(\hat{I}^i,I,\bar{M}^i)$.
        \State $\log\pi_{\theta_{\text{old}}}^{\text{sem},i} \leftarrow \text{Mean}_{k\in M^i}\big[\log\pi_{\theta_{\text{old}}}^{(k)}\big]$,\quad
              $\log\pi_{\theta_{\text{old}}}^{\text{pres},i} \leftarrow \text{Mean}_{k\in \bar{M}^i}\big[\log\pi_{\theta_{\text{old}}}^{(k)}\big]$.
    \EndFor
    \For{\text{each training mini-batch}} \Comment{Policy Update}
        \State $A_{\text{sem}},A_{\text{pres}} \leftarrow \text{GroupNorm}(r_{\text{sem}}, r_{\text{pres}})$.
        \State $\log\pi_\theta^{\text{sem}} \leftarrow \text{Mean}_{k\in M}\big[\log\pi_\theta^{(k)}\big]$,\quad
              $\log\pi_\theta^{\text{pres}} \leftarrow \text{Mean}_{k\in \bar{M}}\big[\log\pi_\theta^{(k)}\big]$.
        \State $\rho_{\text{sem}}=\exp(\log\pi_\theta^{\text{sem}}-\log\pi_{\theta_{\text{old}}}^{\text{sem}})$,\quad
              $\rho_{\text{pres}}=\exp(\log\pi_\theta^{\text{pres}}-\log\pi_{\theta_{\text{old}}}^{\text{pres}})$.
        \State $\mathcal{L}_{\text{sem}}=\text{ClipPG}(A_{\text{sem}},\rho_{\text{sem}})$,\quad
              $\mathcal{L}_{\text{pres}}=\text{ClipPG}(A_{\text{pres}},\rho_{\text{pres}})$.
        \State $\theta \leftarrow \theta - \lambda \nabla_\theta
        \big(\lambda_{\text{sem}}\mathcal{L}_{\text{sem}}
        +\lambda_{\text{pres}}\mathcal{L}_{\text{pres}}
        +\beta\mathcal{L}_{\text{KL}}\big)$.
    \EndFor
    \State $\pi_{\theta_{\text{old}}} \leftarrow \pi_\theta$.
\EndFor
\end{algorithmic}
\textbf{Output:} Optimized policy $\pi_\theta$.
\end{algorithm}

\subsection{Region-Decoupled Policy Optimization}
\label{subsec:optimization-framework}
To mitigate interference between editing and preservation objectives, we further propose a region-decoupled policy gradient optimization framework. The core idea is to apply each reward signal only to the corresponding spatial region, rather than mixing them as one reward and conducting global policy updates. Specifically, during training iterations, we first sample a mini-batch of image-instruction pairs $(I, c)$ and generate a group of $G$ edited samples $\{\hat{I}^i\}_{i=1}^G \sim \pi_{\theta}(\cdot \mid I, c)$ from the current policy. Then, for each sample $\hat{I}^i$, we infer the editing mask $M$ and compute editing reward $r_{\text{sem}}$ and preservation reward $r_{\text{pres}}$, respectively. Instead of combining them into a single scalar, we feed them into group normalization independently to obtain their corresponding advantages:
\begin{equation}
A_{\text{sem}} = \frac{r_{\text{sem}} - \operatorname{mean}\!\left(\{r_{\text{sem}}^{\,i}\}_{i=1}^{G}\right)}{\operatorname{std}\!\left(\{r_{\text{sem}}^{\,i}\}_{i=1}^{G}\right)},\
A_{\text{pres}}=\frac{r_{\text{pres}}-\operatorname{mean}\!\left(\{r_{\text{pres}}^{\,i}\}_{i=1}^{G}\right)}{\operatorname{std}\!\left(\{r_{\text{pres}}^{\,i}\}_{i=1}^{G}\right)}.
\end{equation}
To ensure the editing reward supervises the edit regions while the preservation reward supervises the non-edit regions, we further decouple the policy gradient at token level. Let $\pi_{\theta}^{(k)}$ denote the generation probability of the $k$-th image token, we partition the token set with editing mask $M$ and compute region-aggregated log probabilities:
\begin{equation}
\log \pi_\theta^{\text{sem}}=\frac{1}{|M|}\sum_{k \in M}\log \pi_\theta^{(k)},\
\log \pi_\theta^{\text{pres}}=\frac{1}{|\bar{M}|}\sum_{k \in \bar{M}}\log \pi_\theta^{(k)}.
\end{equation}
Similarly, we define the corresponding log probabilities under the old policy, denoted by $\log \pi_{\theta_{\text{old}}}^{\text{sem}}$ and $\log \pi_{\theta_{\text{old}}}^{\text{pres}}$.
Based on these quantities, we construct two independent optimization objectives:
\begin{equation}
\mathcal{L}_{\text{sem}}=\text{ClipPG}(A_{\text{sem}}, \rho_{\text{sem}}),\
\mathcal{L}_{\text{pres}}=\text{ClipPG}(A_{\text{pres}}, \rho_{\text{pres}}),
\end{equation}
where $\text{ClipPG}(\cdot)$ denotes the GRPO-style clipped policy gradient objective, $\rho_{\text{sem}}=\exp(\log \pi_\theta^{\text{sem}}-\log \pi_{\theta_{\text{old}}}^{\text{sem}})$ and
$\rho_{\text{pres}}=\exp(\log \pi_\theta^{\text{pres}}-\log \pi_{\theta_{\text{old}}}^{\text{pres}})$ are the region-specific policy ratios. Finally, the overall training objective is defined as the weighted combination of the two region-level losses together with a KL regularization term:
\begin{equation}
\mathcal{L}=\lambda_{\text{sem}}\,\mathcal{L}_{\text{sem}}+\lambda_{\text{pres}}\,\mathcal{L}_{\text{pres}} + \beta
\mathcal{L}_\text{KL},
\end{equation}
where $\lambda_{\text{sem}}$ and $\lambda_{\text{pres}}$ balance the relative importance of editing and preservation during optimization. The overall optimization framework is summarized in Algorithm~\ref{alg:edit-grpo}.

\begin{table*}[t]
\centering
\caption{Quantitative comparison on GEdit-Bench~\cite{liu2025step1x}. Best scores are in \textbf{bold}, second-best in \underline{underlined}.}
\label{tab:main_result_gedit}
\resizebox{\linewidth}{!}{
\setlength{\tabcolsep}{14pt}
\begin{tabular}{l|cccc|ccc}
\toprule
\multirow{2}{*}{\textbf{Model}} &\multicolumn{4}{c}{\textbf{Preservation}} & \multicolumn{3}{c}{\textbf{GEdit-Bench-EN}} \\
& UR $\uparrow$ & PSNR $\uparrow$ & SSIM $\uparrow$ & LPIPS $\downarrow$ & G\_SC $\uparrow$ & G\_PQ $\uparrow$ & G\_O $\uparrow$ \\
\midrule
Instruct-Pix2Pix~\cite{brooks2022instructpix2pix} & - & - & - & - &3.58 & 5.49 & 3.68 \\
UniWorld-V1~\cite{Lin2025UniWorldV1HS} & - & - & - & - & 4.93 & 7.43 & 4.85 \\
OmniGen~\cite{xiao2025omnigen} & - & - & - & - & 5.96 & 5.89 & 5.06 \\
\midrule
FLUX.1-Kontext [Dev] & 57.56 & 22.36 & 0.639 & 0.201 & 6.58 & 7.43 & 6.04 \\
+ Edit-R1~\cite{li2025uniworld} & 52.01 & 19.53 & 0.589 & 0.236 & \textbf{7.32} & \underline{7.54} & \textbf{6.84} \\
\rowcolor{gray!18}
\textbf{+ Edit-GRPO(Ours)} & \textbf{60.30} & \textbf{22.91} & \textbf{0.663} & \textbf{0.185} & \underline{6.81} & \textbf{7.56} & \underline{6.34} \\
\midrule
Qwen-Image-Edit [2509] & 56.69 & 21.61 & 0.616 & 0.208 & 8.23 & \textbf{7.98} & 7.70 \\
+ Edit-R1~\cite{li2025uniworld} & 50.97 & 20.53 & 0.597 & 0.229 & \underline{8.34} & 7.88 & \underline{7.77} \\
\rowcolor{gray!18}
\textbf{+ Edit-GRPO(Ours)} & \textbf{63.04} & \textbf{23.07} & \textbf{0.657} & \textbf{0.182} & \textbf{8.40} & \underline{7.94} & \textbf{7.83} \\
\bottomrule
\end{tabular}}
\end{table*}

\begin{table*}[t]
\centering
\caption{Quantitative comparison on ImgEdit~\cite{Ye2025ImgEditAU}. Best scores are in \textbf{bold}, second-best in \underline{underlined}.}
\label{tab:main_results_ImgEdit}
\setlength{\tabcolsep}{1.5pt}
\resizebox{\textwidth}{!}{
\begin{tabular}{l|cccc|cccccccccc}
\toprule
\multirow{2}{*}{\textbf{Model}} &\multicolumn{4}{c}{\textbf{Preservation}} & \multicolumn{10}{c}{\textbf{ImgEdit} } \\
 & UR $\uparrow$ & PSNR $\uparrow$ & SSIM $\uparrow$ & LPIPS $\downarrow$ & Add & Adjust & Extract & Replace & Remove & Background & Style & Hybrid & Action & \textbf{Overall} \\
\midrule
Instruct-Pix2Pix~\cite{brooks2022instructpix2pix} & - & - & - & - & 2.45 & 1.83 & 1.44 & 2.01 & 1.50 & 1.44 & 3.55 & 1.20 & 1.46 & 1.88 \\
OmniGen~\cite{xiao2025omnigen} & - & - & - & - & 3.47 & 3.04 & 1.71 & 2.94 & 2.43 & 3.21 & 4.19 & 2.24 & 3.38 & 2.96 \\
UniWorld-V1~\cite{Lin2025UniWorldV1HS} & - & - & - & - & 3.82 & 3.64 & 2.27 & 3.47 & 3.24 & 2.99 & 4.21 & 2.96 & 2.74 & 3.26 \\
\midrule
FLUX.1 Kontext [Dev] & 56.67 & 22.92 & 0.690 & 0.163 & 3.67 & 3.81 & 2.10 & 4.18 & 3.01 & 3.75 & 4.42 & 2.61 & 3.92 & 3.52 \\
+ Edit-R1~\cite{li2025uniworld} & 45.84 & 18.82 & 0.581 & 0.221 & 3.86 & 3.90 & 2.67 & 4.40 & 3.39 & 4.07 & 4.59 & 3.04 & 4.12 & \textbf{3.82} \\
\rowcolor{gray!18}
\textbf{+ Edit-GRPO(Ours)} & \textbf{59.65} & \textbf{23.25} & \textbf{0.706} & \textbf{0.161} & 3.68 & 3.88 & 2.47 & 4.31 & 3.33 & 3.85 & 4.35 & 2.78 & 3.92 & \underline{3.65} \\
\midrule
Qwen-Image-Edit [2509] & 47.70 & 19.68 & 0.567 & 0.220 & 4.31 & 4.37 & 3.33 & 4.60 & 4.42 & 4.26 & 4.81 & 3.61 & 4.27 & 4.26  \\
+ Edit-R1~\cite{li2025uniworld} & 38.18 & 17.91 & 0.528 & 0.256 & 4.34 & 4.21 & 3.86 & 4.66 & 4.48 & 4.20 & 4.88 & 3.42 & 4.47 & \textbf{4.34} \\
\rowcolor{gray!18}
\textbf{+ Edit-GRPO(Ours)} & \textbf{51.81} & \textbf{20.65} & \textbf{0.608} & \textbf{0.204} & 4.27 & 4.40 & 3.42 & 4.66 & 4.43 & 4.26 & 4.85 & 3.55 & 4.56 & \underline{4.30} \\
\bottomrule
\end{tabular}}
\end{table*}

\section{Experiment}
\label{sec:experiment}
\subsection{Setup}
\label{subsec:Setup}
\textbf{Dataset.} We adopt HumanEdit~\cite{bai2024humanedit} as our training dataset where each sample consists of an input image and a natural language editing instruction describing the expected modification. The dataset covers diverse editing types, including object addition, removal, replacement, action modification, and counting-based edits, providing a broad range of editing scenarios.

\textbf{Implementation Details.} We use FLUX.1-Kontext [Dev]~\cite{labs2025flux} and Qwen-Image-Edit [2509]~\cite{wu2025qwen} as our base models. For editing mask generation, we employ Qwen2.5-14B-Instruct~\cite{Yang2024Qwen25TR} as the object prompt extractor and SAM3~\cite{carion2025sam3segmentconcepts} as the segmentation model. We employ Qwen3-VL-32B-Instruct~\cite{Qwen3-VL} as the MLLM-based score evaluator. 
More implementation details are provided in the Appendix~\ref{appen:Implementation Details}.

\textbf{Semantic Editing Evaluation.} 
For quantitative semantic editing evaluation, we employ two representative benchmarks: GEdit-Bench~\cite{liu2025step1x}, which evaluates image editing through diverse language instructions, and ImgEdit~\cite{Ye2025ImgEditAU}, which unifies multiple tasks into a common comparison framework.

\textbf{Locality Preservation Evaluation.} We introduce several locality-aware metrics to quantify locality preservation. First, we introduce \textbf{Unchanged Ratio (UR)}, defined as the proportion of pixels in non-edit regions whose absolute difference falls below a small threshold before and after editing. In addition, we further report PSNR, SSIM and LPIPS to measure low-level fidelity, structural consistency, and perceptual similarity, respectively. More evaluation details are provided in  Appendix~\ref{appen:Evaluation Details}.

\begin{figure*}[t]
\centering
\includegraphics[width=\linewidth]{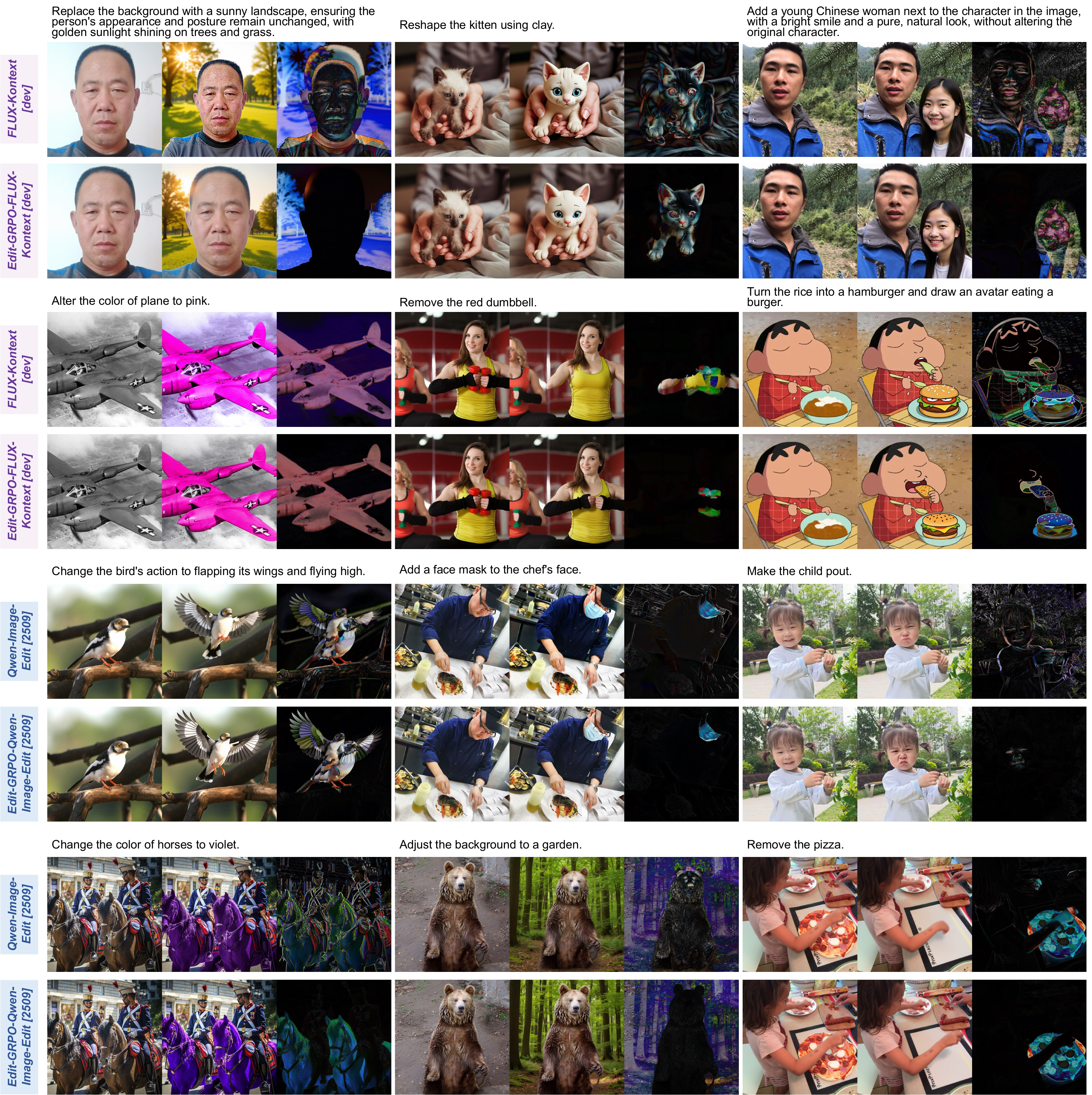}
\caption{Qualitative comparison of editing capabilities before and after policy optimization. Our approach demonstrates superior semantic editing and locality preservation over the base models.}
\label{fig:samples}
\vspace{-25pt}
\end{figure*}

\subsection{Main Results}
\label{subsec:main_results}
\begin{wrapfigure}{r}{0.40\linewidth}
\vspace{-5pt}
\centering
\includegraphics[width=\linewidth]{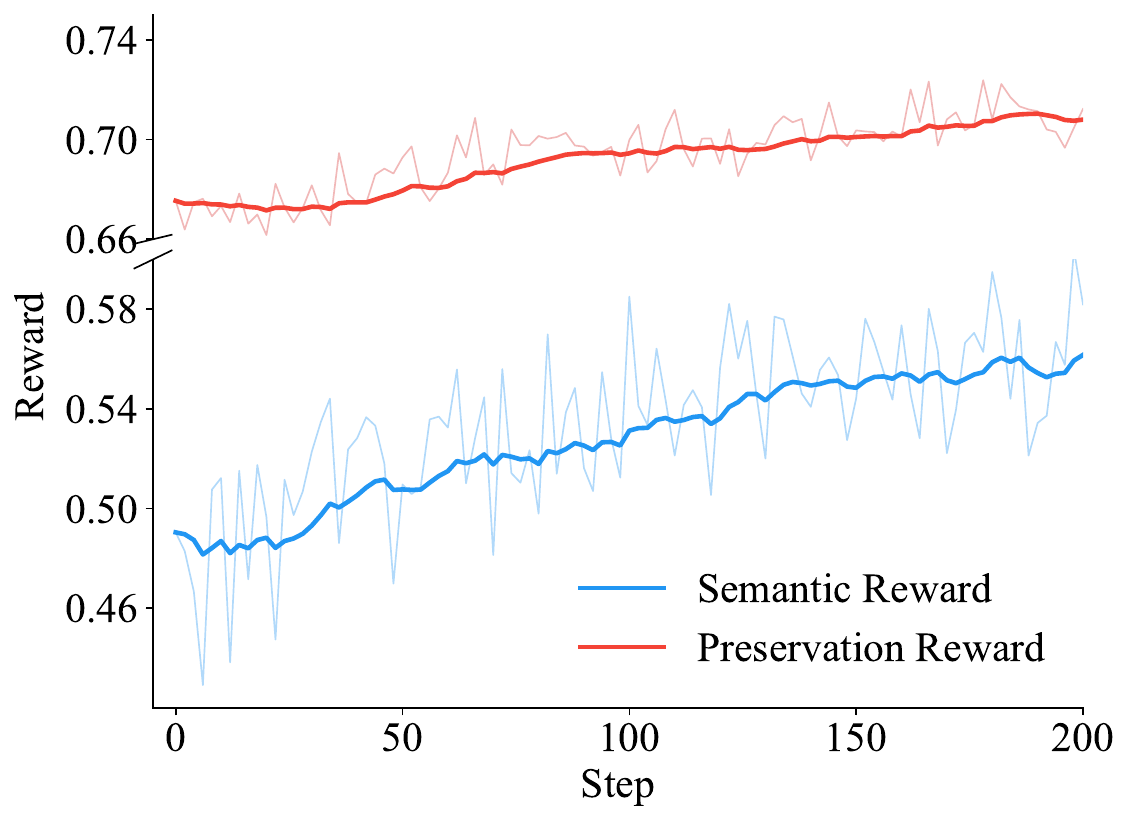}
\vspace{-15pt}
\caption{Training reward curves on FLUX.1-Kontext [Dev].}
\label{fig:reward_curve}
\vspace{-10pt}
\end{wrapfigure}

We evaluate Edit-GRPO on both semantic editing and locality preservation. Quantitative results are reported in Table~\ref{tab:main_result_gedit} and Table~\ref{tab:main_results_ImgEdit}. Qualitative comparisons are shown in Figure~\ref{fig:samples}. We further visualize the smoothed reward trajectory during training in Figure~\ref{fig:reward_curve}.

\textbf{Global reward optimization improves editing scores at the expense of locality preservation.}
As shown in Table~\ref{tab:main_result_gedit} and Table~\ref{tab:main_results_ImgEdit}, the global optimization method Edit-R1~\cite{li2025uniworld} improves semantic editing scores while substantially harming locality preservation. On GEdit-Bench, applying Edit-R1 to FLUX.1-Kontext [Dev] increases G\_O to 6.84, but causes clear degradation in all preservation metrics (UR: 57.56\% $\rightarrow$ 52.01\%, PSNR: 22.36 $\rightarrow$ 19.53, SSIM: 0.639 $\rightarrow$ 0.589, LPIPS: 0.201 $\rightarrow$ 0.236). Similar experimental results are observed for Qwen-Image-Edit [2509] and on the ImgEdit benchmark. These results indicate that global optimization tends to encourage aggressive edits, introducing substantial content modifications outside the target regions.

\textbf{Edit-GRPO consistently improves locality preservation while retaining competitive editing performance.}
On GEdit-Bench (Table~\ref{tab:main_result_gedit}), Edit-GRPO increases UR from 57.56\% to 60.30\% for FLUX.1-Kontext [Dev] and from 56.69\% to 63.04\% for Qwen-Image-Edit [2509]. Meanwhile, semantic editing performance is preserved or improved relative to the base models, with G\_O increasing from 6.04 to 6.34 for FLUX.1-Kontext [Dev] and from 7.70 to 7.83 for Qwen-Image-Edit [2509]. Similar trends are observed on ImgEdit (Table~\ref{tab:main_results_ImgEdit}), where Edit-GRPO improves all preservation metrics while maintaining competitive overall editing scores. These results demonstrate that Edit-GRPO effectively improves locality preservation while retaining competitive editing performance, mitigating the preservation-editing trade-off in prior approaches.


\textbf{Qualitative results.}
As illustrated in Figure~\ref{fig:samples}, Edit-GRPO produces edits that are more localized to the target regions and better preserve surrounding content compared to the base models, further demonstrating that our method can improve locality preservation while maintaining editing precision.

\subsection{Ablation Study}
In this section, we conduct a systematic analysis to evaluate the contribution of each component in Edit-GRPO. All experiments are trained on FLUX.1-Kontext [Dev] and evaluated on GEdit-Bench.

\textbf{Decoupled Optimization vs. Global Optimization.}
We first study different objective designs. As shown in Table~\ref{tab:ablation_edit_and_nonedit_losses}, Semantic-Only optimization achieves the highest editing score (G\_O 6.44) but substantially degrades locality preservation (UR: 57.56\% $\rightarrow$ 52.54\%), while Preservation-Only optimization yields the best preservation metrics (UR 64.10\%) but significantly hurts editing performance (G\_O: 6.04 $\rightarrow$ 5.09). This phenomenon indicates a clear conflict between the two objectives when optimized alone. In contrast, our decoupled optimization improves preservation metrics (UR 60.30\%) while maintaining a strong editing score (G\_O 6.34), enabling simultaneous optimization of semantic editing and locality preservation.
\begin{table*}[t]
\centering
\caption{Ablation on optimization objectives. Best scores are in \textbf{bold}, second-best in \underline{underlined}.}
\label{tab:ablation_edit_and_nonedit_losses}
\resizebox{\linewidth}{!}{
\setlength{\tabcolsep}{16pt}
\begin{tabular}{l|cccc|ccc}
\toprule
\textbf{Loss Design} & UR $\uparrow$ & PSNR $\uparrow$ & SSIM $\uparrow$ & LPIPS $\downarrow$ & G\_SC $\uparrow$ & G\_PQ $\uparrow$ & G\_O $\uparrow$ \\
\midrule
FLUX.1-Kontext [Dev] & 57.56 & 22.36 & 0.639 & 0.201 & 6.58 & 7.43 & 6.04 \\
\hline
Semantic-Only & 52.54 & 21.23 & 0.624 & 0.214 & \textbf{7.02} & 7.29 & \textbf{6.44} \\
Preservation-Only & \textbf{64.10} & \textbf{24.49} & \textbf{0.700} & \textbf{0.176} & 5.42 & \textbf{7.76} & 5.09 \\
Combined Optimization & 57.90 & 22.80 & 0.658 & 0.191 & 6.63 & 7.54 & 6.15 \\
\rowcolor{gray!18}
Decoupled Optimization & \underline{60.30} & \underline{22.91} & \underline{0.663} & \underline{0.185} & \underline{6.81} & \underline{7.56} & \underline{6.34} \\
\bottomrule
\end{tabular}}
\vspace{-3pt}
\end{table*}
Notably, while mixing the two rewards into a scalar for global optimization yields slight improvements (G\_O: 6.04 $\rightarrow$ 6.15, UR: 57.56\% $\rightarrow$ 57.90\%), decoupled optimization achieves stronger improvements (G\_O: 6.34, UR: 60.30\%), demonstrating the effectiveness of the proposed decoupled optimization framework.

\begin{table*}[t]
    \begin{minipage}[t]{0.48\linewidth}
        \centering
        \caption{Ablation on semantic reward designs.}
        \label{tab:ablation_edit_loss}
        \resizebox{\linewidth}{!}{
        \setlength{\tabcolsep}{4.5pt}
        \begin{tabular}{l|ccc}
        \toprule
        \textbf{Reward Design} & G\_SC $\uparrow$ & G\_PQ $\uparrow$ & G\_O $\uparrow$ \\
        \midrule  
        FLUX.1-Kontext [Dev] & 6.58 & 7.43 & 6.04 \\
        \hline
        Preservation-Only & 5.42 & \textbf{7.76} & 5.09 \\
        + CLIP reward & 6.35 & 7.50 & 5.94  \\
         \quad+ MLLM (discrete) reward & 6.67 & 7.39 & 6.14  \\
        \rowcolor{gray!18}
         \quad+ MLLM (logit) reward & \textbf{6.81} & 7.56 & \textbf{6.34} \\
        \bottomrule
        \end{tabular}
        }
    \end{minipage}
    \hfill 
    \begin{minipage}[t]{0.50\linewidth}
        \centering
        \caption{Ablation on preservation reward designs.}
        \label{tab:ablation_non_edit_loss}
        \resizebox{\linewidth}{!}{
        \setlength{\tabcolsep}{4.5pt}
        \begin{tabular}{l|cccc}
        \toprule
        \textbf{Reward Design} & UR $\uparrow$ & PSNR $\uparrow$ & SSIM $\uparrow$ & LPIPS $\downarrow$  \\
        \midrule
        FLUX.1-Kontext [Dev] & 57.56 & 22.36 & 0.639 & 0.201 \\
        \hline
        Semantic-Only & 52.54 & 21.23 & 0.624 & 0.214 \\
        +LPIPS reward & 53.09 & 21.66 & 0.638 & 0.202 \\
        \quad+SSIM reward & 53.47 & 22.08 & 0.653 & 0.192 \\
        \rowcolor{gray!18}
        \quad\quad+ L1 reward & \textbf{60.30} & \textbf{22.91} & \textbf{0.663} & \textbf{0.185} \\
        \bottomrule
        \end{tabular}}
    \end{minipage}
\vspace{-10pt}
\end{table*}

\textbf{Semantic Reward Design.}
We ablate different semantic reward designs in Table~\ref{tab:ablation_edit_loss}. Without semantic reward, the model suffers from a clear performance degradation in semantic following (G\_O: 6.04 $\rightarrow$ 5.09). Adding the CLIP-based directional reward improves G\_O to 5.94 and incorporating the MLLM-based discrete reward further improves G\_O to 6.14. Replacing the MLLM-based discrete reward with logit reward further improves G\_O from 6.14 to 6.34, suggesting that the logit reward can provide more fine-grained signals for instruction-compliant editing.

\textbf{Preservation Reward Design.}
We further ablate preservation reward design in Table~\ref{tab:ablation_non_edit_loss}. Without preservation reward, locality preservation degrades substantially compared to the base model (UR: 57.56\% $\rightarrow$ 52.54\%). Adding the LPIPS reward improves perceptual preservation, reducing LPIPS from 0.214 to 0.202. Further incorporating the SSIM reward strengthens structural consistency, achieving SSIM of 0.653. Finally, adding the pixel-level L1 reward achieves the largest improvement in locality preservation, increasing UR from 53.47\% to 60.30\%. These results suggest that perceptual, structural, and pixel-level rewards provide complementary preservation signals, with the L1 reward playing a crucial role in suppressing unintended modifications in non-edit regions.


\section{Conclusion}
In this paper, we present Edit-GRPO, a general locality-preserving policy optimization framework for image editing. By explicitly aligning optimization signals with the spatial structure of editing tasks, Edit-GRPO enables targeted improvements in editing quality while effectively preserving surrounding context. Extensive experiments demonstrate that region-decoupled optimization is a principled and effective solution to the long-standing locality challenge in image editing. We hope this work inspires future research on spatially aware optimization strategies for generative models.

{
\small
\bibliographystyle{plainnat} \bibliography{neurips_2026}
}

\clearpage
\newpage

\appendix

\section{Implementation Details}
\label{appen:Implementation Details}
\subsection{Training Details}
\label{appen:Training Details}
In our implementation, we use FLUX.1-Kontext [Dev]~\cite{labs2025flux} and Qwen-Image-Edit [2509]~\cite{wu2025qwen} as our base models. For editing mask generation, we employ Qwen2.5-14B-Instruct~\cite{Yang2024Qwen25TR} as the object prompt extractor and SAM3~\cite{carion2025sam3segmentconcepts} as the segmentation model. We employ Qwen3-VL-32B-Instruct~\cite{Qwen3-VL} as the MLLM-based score evaluator for the edited images. All images are resized to a resolution of $512 \times 512$ during training. To improve memory efficiency, we adopt Fully Sharded Data Parallelism (FSDP) for the text encoder and enable gradient checkpointing when training Qwen-Image-Edit [2509]. All training experiments are conducted in \texttt{bfloat16} mixed precision.

We use the AdamW~\cite{loshchilov2017decoupled} optimizer with a learning rate of $3 \times 10^{-4}$, $\beta_1 = 0.9$, $\beta_2 = 0.999$, and a weight decay of $10^{-4}$. We apply a KL regularization term with coefficient $\beta = 1 \times 10^{-4}$ to constrain the policy from deviating excessively from the reference model. We maintain an exponential moving average (EMA) of model parameters with decay 0.9. We adopt per-prompt normalization with global standard deviation across the group to reduce variance. For the reward function, we set the semantic reward weights to $\lambda_{\text{vlm}}=4.0$ and $\lambda_{\text{clip}}=1.0$, and the preservation reward weights to $\lambda_{\text{diff}}=4.0$, $\lambda_{\text{ssim}}=2.0$, and $\lambda_{\text{lpips}}=1.0$. The region-decoupled loss weights are set to $\lambda_{\text{sem}}=0.5$ for the semantic branch and $\lambda_{\text{pres}}=1.0$ for the preservation branch.

\subsection{Evaluation Details}
\label{appen:Evaluation Details}
The evaluation experiments consists of two distinct components: semantic editing evaluation and locality preservation evaluation. For semantic editing evaluation, we strictly follow the GEdit-Bench~\cite{liu2025step1x} and the ImgEdit~\cite{Ye2025ImgEditAU} benchmarks, using GPT4.1 for the semantic score evaluation. For locality preservation evaluation, we design an auto mask-based evaluation pipeline. Specifically, we first use Qwen2.5-14B-Instruct~\cite{Yang2024Qwen25TR} to parse the editing instruction and extract structured descriptions of the target content. These descriptions and images are then fed into the SAM3~\cite{carion2025sam3segmentconcepts} model to generate high-resolution masks for the edit target. To acquire robust segmentation in diverse editing scenarios, we apply region decoupling to both the original image and the edited image, obtaining two editing masks accordingly. The final segmentation mask is computed as the union of the two masks. All the remaining areas are treated as the non-edit regions. 

Based on the region segmentation masks, we compute the following locality preservation metrics exclusively in non-edit regions:

\textbf{Unchanged Ratio (UR).}
This metric measures the proportion of pixels in non-edit regions that remain nearly unchanged after editing. For each pixel $p$ in non-edit regions $\bar{M}$, we compute the L1 distance across RGB channels between the original image $I$ and the edited image $\hat{I}$. A pixel is considered unchanged if the distance falls below a threshold $\tau$. In our implementation, we set $\tau = 20$ in RGB pixel space. Formally, the UR metric is defined as:
\begin{equation}
\mathrm{UR} = \frac{\left|\left\{p \in \bar{M} : \| I(p) - \hat{I}(p) \|_{1} < \tau \right\}\right|}{|\bar{M}|}.
\end{equation}

\textbf{PSNR.}
Peak Signal-to-Noise Ratio measures pixel-level reconstruction fidelity in non-edit regions. We compute the mean squared error (MSE) between the original image $I$ and the edited image $\hat{I}$ over pixels in the non-edit regions $\bar{M}$ across all RGB channels:
\begin{equation}
\mathrm{PSNR} = 10 \log_{10} \left( \frac{255^2}{\mathrm{MSE}} \right),
\end{equation}
where $\mathrm{MSE} = \frac{1}{3|\bar{M}|} \sum_{p \in \bar{M}} \| I(p) - \hat{I}(p) \|_2^2$.

\textbf{SSIM.}
Structural Similarity Index (SSIM) evaluates the structural consistency between the original and edited images. Following the standard protocol, we convert both images to grayscale and compute SSIM within the non-edit regions $\bar{M}$.

\textbf{LPIPS.} Learned Perceptual Image Patch Similarity (LPIPS) measures perceptual similarity using deep visual features. Following the standard LPIPS implementation, images are first normalized to the range $[-1,1]$ and then passed through a pretrained VGG-based network~\cite{zhang2018unreasonable}. The perceptual distance is computed in non-edit regions $\bar{M}$, where lower LPIPS values indicate better perceptual preservation.

In practice, when conducting locality preservation evaluation, we explicitly \textbf{exclude} samples whose task type belongs to \textit{style\_change}, \textit{text\_change} and \textit{tone\_transfer} as these editing categories involve global or diffuse transformations where SAM3~\cite{carion2025sam3segmentconcepts} cannot accurately segment a well-defined local edit region, making the mask-based locality evaluation unreliable. We further apply two filtering criteria to handle cases where the generated segmentation masks are inaccurate: (i) samples where the non-edit regions cover less than 5\% of the image are discarded, as they likely represent near-global edits for which mask-based evaluation is not meaningful; (ii) samples where the edit regions cover less than 1\% of the image are also discarded, as this typically indicates that SAM3 fails to localize the edited object and generate a meaningful segmentation mask.

\subsection{Hyperparameters and Computation Resources}
\label{appen:Hyperparameters}
We report the key hyperparameters of Edit-GRPO in Table~\ref{tab:hyperparameters}. All training and evaluation experiments are conducted on a single machine with 8 NVIDIA H20 96 GB GPUs.

\begin{table*}[t]
\centering
\caption{Hyperparameter settings for Edit-GRPO. We report the configurations for both FLUX.1-Kontext [Dev] and Qwen-Image-Edit [2509] backbones.}
\label{tab:hyperparameters}
\resizebox{\linewidth}{!}{
\setlength{\tabcolsep}{20pt}
\begin{tabular}{lcc}
\toprule
\textbf{Hyperparameter} & \textbf{FLUX.1-Kontext [Dev]} & \textbf{Qwen-Image-Edit [2509]} \\
\midrule
\multicolumn{3}{c}{\textit{Training Settings}} \\
\midrule
Batch Size (per device) & $1$ & $1$ \\
Gradient Accumulation Steps & $24$ & $24$ \\
Learning Rate & $3 \times 10^{-4}$ & $3 \times 10^{-4}$ \\
Adam ($\beta_1$, $\beta_2$) & $0.9, 0.999$ & $0.9, 0.999$ \\
Weight Decay & $1 \times 10^{-4}$ & $1 \times 10^{-4}$ \\
EMA Decay & $0.9$ & - \\
\midrule
\multicolumn{3}{c}{\textit{Sampling Settings}} \\
\midrule
Image Resolution & $512 \times 512$ & $512 \times 512$ \\
Sampling Inference Steps & $6$ & $10$ \\
Images Per Prompt & $8$ & $8$ \\
Noise Level & 0.9 & 0.7 \\
\midrule
\multicolumn{3}{c}{\textit{Reward Weights}} \\
\midrule
MLLM Semantic Score & $4.0$ & $4.0$ \\
CLIP Directional Score & $1.0$ & $1.0$ \\
L1 Preservation Score & $4.0$ & $4.0$ \\
SSIM Preservation Score & $2.0$ & $2.0$ \\
LPIPS Preservation Score & $1.0$ & $1.0$ \\
\midrule
\multicolumn{3}{c}{\textit{Loss Coefficients}} \\
\midrule
Semantic Loss  & $0.5$ & $0.5$ \\
Preservation Loss  & $1.0$ & $1.0$ \\
KL Divergence ($\beta$) & $1 \times 10^{-4}$ & $1 \times 10^{-4}$ \\
\bottomrule
\end{tabular}}
\end{table*}


\section{Prompt Template}
\label{appen:Prompt Template}
We provide the prompt template used for MLLM-based semantic score evaluation in Figure~\ref{fig:template}.

\begin{figure}[t]
\small
\fbox{
\begin{minipage}{0.90\linewidth}
\textbf{Template for MLLM-based Semantic Score Evaluation:}

You are an image editing evaluator. Evaluate how well the edited image follows the instruction.

\textbf{Original Image}: [First image] \\
\textbf{Edited Image}: [Second image] \\
\textbf{Instruction}: \{instruction\} \\
You need to rate the editing result from 0 to 5 based on the accuracy and quality of the edit.

\textbf{Scoring Rules:}

0: The wrong object was edited, or the edit completely fails to meet the requirements.  

5: The correct object was edited, the requirements were met, and the visual result is high quality.

\textbf{Response Format:}

Directly output the score number (0--5).

\end{minipage}
}
\caption{Prompt template used for MLLM-based semantic score evaluation.}
\label{fig:template}
\end{figure}

\section{Limitations}
\label{appen:limitaions}
Despite the improvements achieved by our method, several limitations remain to be addressed. First, the effectiveness of our Edit-GRPO relies on the quality of the segmentation masks used to separate edit and non-edit areas. However, in practice, the quality of automatically generated masks can vary depending on the complexity of the editing scenario. Imperfect masks may lead to inaccurate region assignments, which can affect the stability and effectiveness of the optimization process.

Second, although our region-decoupled optimization framework can alleviate the conflict between semantic editing and locality preservation objectives, this conflict is not eliminated completely. Future work could explore more advanced collaborative optimization strategies to better balance editing strength and background preservation, further improving the robustness of instruction-based image editing systems.

\section{Societal Impacts}
\label{appen:Impacts}
\subsection{Positive Impacts}
The proposed Edit-GRPO framework improves the controllability and reliability of instruction-based image editing by explicitly optimizing both semantic editing and locality preservation without requiring additional manual annotations. By decoupling editing and preservation objectives during policy optimization, our method encourages models to perform targeted modifications while maintaining the surrounding visual context. This improvement can benefit a range of applications such as digital content creation, photo retouching, graphic design, and e-commerce visualization, where localized edits are often required while the rest of the image should remain unchanged. Beyond professional applications, improved locality-aware editing can also make visual editing tools more accessible to non-expert users. With natural language instructions, users can perform targeted edits without extensive experience in professional editing software, potentially lowering the barrier to creative workflows and improving productivity. In addition, explicitly optimizing preservation objectives promotes more predictable editing behavior, which helps reduce unintended visual artifacts and improve user trust in AI-assisted editing systems.

\subsection{Negative Impacts}
Despite these benefits, the proposed method may also introduce potential societal risks. More precise and realistic image editing capabilities could be misused to manipulate visual content in subtle ways while preserving most of the original scene. Such edits may become harder to detect and could facilitate the creation of misleading or deceptive visual media, including manipulated photographs or visual disinformation. Furthermore, more precise and realistic image editing may also be used to modify personal or sensitive images, raising privacy and ethical concerns.

\section{Declaration of LLM usage}
\label{appen:llm-usage}
In this work, LLMs were employed for language-related tasks, including grammar correction, spelling checks, and word choice refinement, to improve the writing fluency of the paper. All scientific content, analyses, and conclusions were conceived, validated, and interpreted by the authors.

\section{More Visualization Results}
\begin{figure*}[t]
\centering
\includegraphics[width=\linewidth]{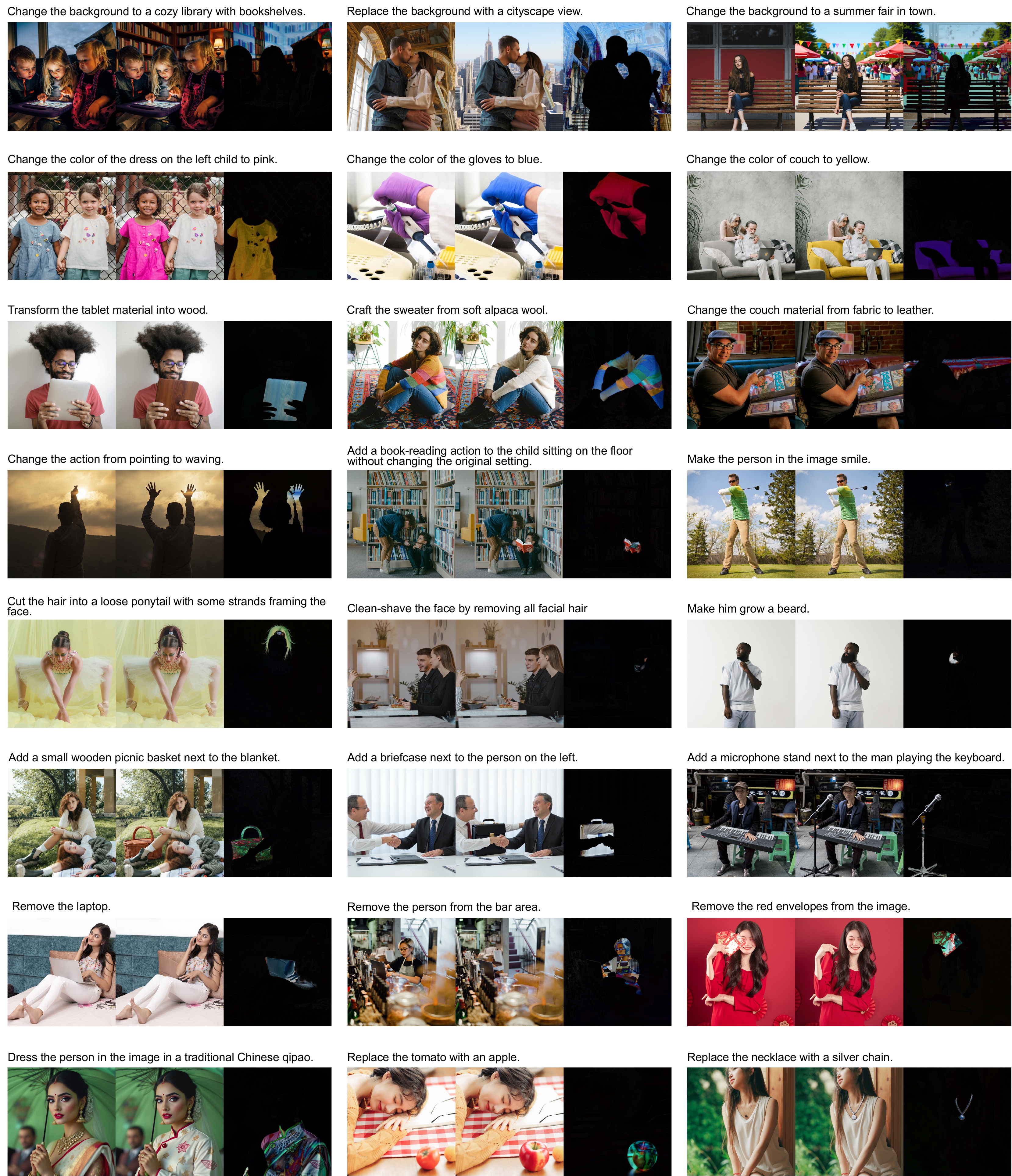}
\caption{The visualization results of Edit-GRPO-FLUX-Kontext [dev].}
\label{fig:more_cases_1}
\end{figure*}

\begin{figure*}[t]
\centering
\includegraphics[width=\linewidth]{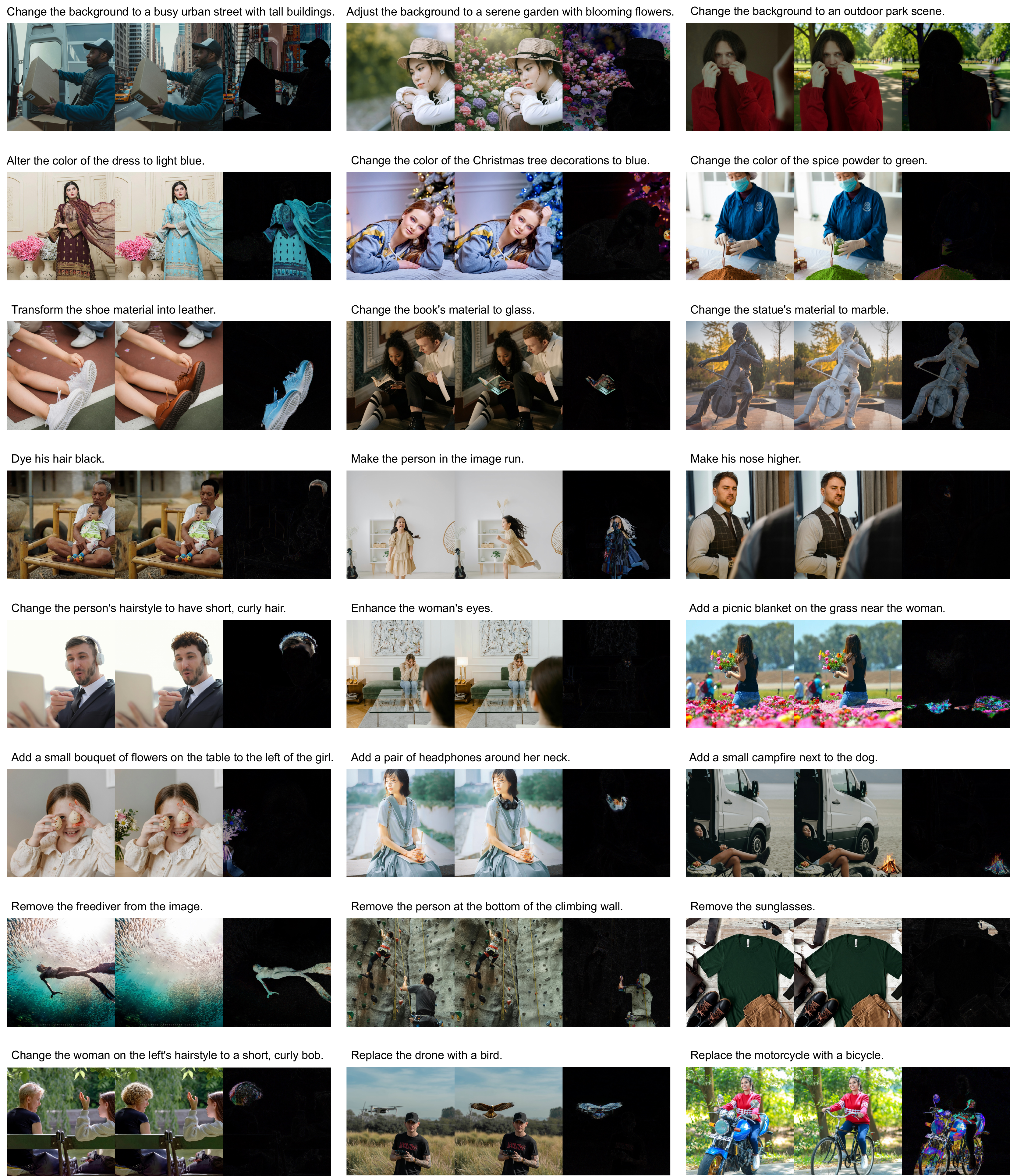}
\caption{The visualization results of Edit-GRPO-Qwen-Image-Edit [2509].}
\label{fig:more_cases_2}
\end{figure*}

We present more visualization results of Edit-GRPO-FLUX-Kontext [dev] and Edit-GRPO-Qwen-Image-Edit [2509] in Figure~\ref{fig:more_cases_1} and Figure~\ref{fig:more_cases_2}.

\end{document}